\documentclass[conference]{IEEEtran}
\IEEEoverridecommandlockouts
\usepackage{cite}
\usepackage{amsmath,amssymb,amsfonts}
\usepackage{algorithmic}
\usepackage{graphicx}
\usepackage{textcomp}
\usepackage{xcolor}
\def\BibTeX{{\rm B\kern-.05em{\sc i\kern-.025em b}\kern-.08em
    T\kern-.1667em\lower.7ex\hbox{E}\kern-.125emX}}
\begin{document}

\title{Simulators for Mobile Social Robots:
State-of-the-Art and Challenges\\
}

\author{\IEEEauthorblockN{Prabhjot Kaur\IEEEauthorrefmark{1},
Zichuan Liu\IEEEauthorrefmark{2}, and
Weisong Shi\IEEEauthorrefmark{3}}
\IEEEauthorblockA{\textit{Department of Computer Science},
\textit{Wayne State University}\\
Detroit, Michigan 48202\\
Email: \IEEEauthorrefmark{1}prabhjotkaur@wayne.edu,
\IEEEauthorrefmark{2}zichuanliu@wayne.edu,
\IEEEauthorrefmark{3}weisong@wayne.edu}}
\maketitle

\begin{abstract}
The future robots are expected to work in a shared physical space with humans \cite{julie}, however, the presence of humans leads to a dynamic environment that is challenging for mobile robots to navigate. The path planning algorithms designed to navigate a collision free path in complex human environments are often tested in real environments due to the lack of simulation frameworks. This paper identifies key requirements for an ideal simulator for this task, evaluates existing simulation frameworks and most importantly, it identifies the challenges and limitations of the existing simulation techniques.

First and foremost, we recognize that the simulators needed for the purpose of testing mobile robots designed for human environments are unique as they must model realistic pedestrian behavior in addition to the modelling of mobile robots. Our study finds that Pedsim\_ros \cite{pedsimros} and a more recent SocNavBench framework \cite{biswas2021socnavbench} are the only two 3D simulation frameworks that meet most of the key requirements defined in our paper. In summary, we identify the need for developing more simulators that offer an ability to create realistic 3D pedestrian rich virtual environments along with the flexibility of designing complex robots and their sensor models from scratch.
\end{abstract}


\section{Introduction}
The social Autonomous Mobile Robots (AMRs) include robots that can perceive their environment, plan and execute a collision free trajectory to reach their destination autonomously in complex human environments \cite{fries}. The current forecast by the International Federation of Robotics (IFR) estimates the market of mobile robots to grow on an average by 97\% annually. Their presence has already increased in response to COVID-19, with public safety and clinical care being the two dominant applications in 2020 \cite{murphy2020applications}. The other active applications include usage in hotels \cite{choi2020service}, \cite{kim2021preference}, factories \cite{schneier2015literature}, construction \cite{construction} and hospitals \cite{health}. Further, there has been an increased interest in using robots to provide quality care for the elderly, either in their own homes \cite{robinson2014role} or in the nursing homes \cite{aloulou2013deployment} to meet the staggering shortage of care givers \cite{miller2017future}. With the current state of the technology, the AMRs are able to achieve some level of autonomy in controlled environments, however, the real world is dynamic and unpredictable which makes it challenging to scale up their deployment in human environments.

This is where strategic testing plays an important role. Since the real world testing is costly and unsafe, it can be complemented with simulation testing which is cost effective, safer and reproducible. The simulation testing in the field of mobile social robots is currently inadequate because of the challenges in recreating realistic 3D scenarios and accurate modelling of human behavior in simulation \cite{biswas2021socnavbench}. Considering the diverse applications of robotics mentioned before, one would expect a simulator to be able to model both indoor environments (such as hospitals, nursing homes, etc.) and outdoor environments. Further, it must model different types of pedestrian behavior including people on wheel chairs, with canes, crowds, etc. to make the virtual environment representative of a real world. Luckily, the study of social behavior of humans is a well-researched topic in psychology \cite{wang}, which has lead to the development of various mathematical models for pedestrian behavior \cite{pedmodels}, \cite{caramuta2017survey}. These models generalize human behavior and they aim to predict pedestrians' next move. Some of these pedestrian models have made their way into robotics where they are integrated with the path planning algorithms to help the AMRs execute a path without colliding with humans. Our survey shows that there is a lack of open-source simulators suitable for testing such algorithms  \cite{pedmodels}, \cite{aroor2018mengeros}. 

So far, there are two 3D frameworks namely, Pedsim\_ros and SocNavBench that have been released as open source to fill the gap. However, these frameworks have their own limitations which are discussed below. In this paper, we first identify key requirements for building an effective simulator for testing AMRs in pedestrian environments. Secondly, we study and evaluate existing simulation techniques from the perspective of simulating robots and pedestrians in a shared environment. Thirdly, we provide our analysis of the current simulation frameworks and discuss the challenges with using existing techniques.

This paper is organized into various sections. Section \ref{motivation} provides background on why a simulator for testing AMRs in a human rich environment is distinct from a standalone robotic simulator or a pedestrian crowd simulator. Section \ref{requirements} identifies requirements for building a holistic simulator. Section \ref{sims} reviews and summarizes the current state-of-the art simulators. Section V provides a thorough evaluation of these simulators through various experiments that we conducted. Section VI discusses challenges that remain to be addressed when it comes to building a holistic simulator for AMR testing in a human rich environment. Section VII reviews related works. Finally, this paper concludes in Section VIII.

\hfill 
 
\hfill 
\section{Motivation and Background}
\label{motivation}
There are many robotic and pedestrian crowd simulators available today that are intended for testing robotic systems and pedestrian/human models, respectively. However, to test novel navigation algorithms for AMRs in pedestrian rich environments, there is a need for a holistic simulator that combines the capabilities of these two types of simulators. Currently, to test these algorithms, the researchers rely on either testing directly in real environments as in \cite{shi2013model} and \cite{shiomi2014towards} or spend significant time in building their own simulation environments as in \cite{zhi2021anticipatory} and \cite{chebotareva2021laser}. Therefore, there is a clear need to have more open source simulators that are available for researchers to use for testing mobile robots in human rich environments. As mentioned before, the simulator for such a testing environment is unique because one must model not just the 3D virtual environment and a robot, but also the models for pedestrians that are behaving realistically. In the next section, we define key requirements that an ideal simulator for this purpose must meet.

\section{Requirements}
\label{requirements}
While testing the AMR, there are many aspects of its performance that must be tested, such as its perception functionality, its interaction with humans, its path planning and control \cite{biswas2021socnavbench}. However, before we can even consider evaluating these capabilities in simulation, one must construct a realistic 3D virtual environment that includes a scene, a model of a robot including its sensors and models of humans/pedestrians. Therefore, we start with identifying \textbf{key requirements that are essential to construct such an environment.} These requirements can then be used as a metric to compare which simulators are best suited for the task at hand. 

\begin{enumerate}
    \item \textbf{3D virtual environment modelling}: The simulator must support the creation of any 3D virtual scene (both indoor and outdoor) such as a nursing home, hospital, trade show set up, stadium, school campus, etc. in which to test and deploy the mobile robot. 
    \item \textbf{Mobile robot modelling}: The simulator must provide the flexibility of modelling a mobile robot from scratch so that every aspect of the robot can be tested thoroughly, such as its on-board sensors, sensor placement, and its physical properties.
    \item \textbf{Sensor modelling}: The sensor suit for any mobile robot is highly dependent on the navigation algorithm being used to guide a robot's path. Some of the common navigation algorithms use visual, ultrasonic, and laser sensors \cite{siegwart2011introduction}. Since each type of physical sensor has its own intrinsic properties such as its sensing range, noise and latency, the simulator must allow one to model a realistic sensor for testing the social mobile robot.
    \item \textbf{Human behavior modelling}: Most importantly, the simulator must realistically model virtual humans in the 3D environment. Therefore, the simulator must have built-in models for modelling human behavior. This requirement is what makes such simulators unique and challenging to construct.
    \item \textbf{Integration with a robotic middleware}: The simulator must support easy integration with a robotic middleware. A middleware is a software framework that resides between the operating system of the host machine and the application layer. It handles lower level essential tasks such as cross communication between various components of a complex system.  Since robotic systems are inherently complex, relying on a middleware to handle low level tasks helps. As an example, the underlying processes involved with AMRs are sensing, motion planning and control. These processes need to cross communicate to achieve the goal of collision avoidance. The number of such processes grows exponentially with the complexity of the task and the number of robots in the scene. 
\end{enumerate}

Besides the requirements identified above, it is important that the simulator is open source, scalable, modular and portable. As part of our review, various simulators were studied while keeping the above requirements in mind. Our survey shows that most traditional simulators are either specialized robot simulators or are specialized pedestrian/crowd simulators \cite{aroor2018mengeros}. However, there are very few holistic simulators that are suitable for testing mobile social robots. The details of these simulation frameworks are discussed in the following section. 

\section{Simulators}
\label{sims}
This section is divided into four subsections. Each subsection reviews simulation frameworks specializing in a certain task as identified in the key requirements in Section  \ref{requirements}.
\subsection{Simulators for 3D environment, robot and sensor modelling}
\label{3D_env_modelling}
The first and foremost requirement to test mobile robots is having a virtual 3D environment and a 3D model of the robot including its on-board sensors. Some of the key characteristics for such simulators are that they must support automatic 3D scene generation and must be open-source with wide community support. Additionally, these simulators must be easy to integrate with robotic middleware systems. This subsection describes various 3D simulators that enable creation of virtual 3D environment, robot model and its sensors..

\subsubsection{Gazebo/Ignition} It is one of the most popular multi-robot 3D simulator \cite{koenig2004design}. It enables the users to generate both indoor and outdoor environments, along with the support for modelling almost all kinds of robots such as humanoid \cite{ivaldi2014tools}, mobile, underwater, aerial, etc.

The 3D scene in Gazebo environment is called a world which consists of several static and dynamic objects called models. The world is defined by the user in a Simulation Description File (SDF) format \cite{sdf}. Further, the users are empowered to customize every detail of the model such as its inertial, visual or collision properties. All aspects of a 3D scene can be controlled through plugins. \cite{prius} showcases the advanced capabilities of Gazebo where the user defines a prius model of a car moving in a simulated M-City. These realistic simulations are possible in Gazebo because of the three main  libraries namely, physics, rendering, and communication libraries \cite{kaur2021survey}. Gazebo is the most popular simulator in the robotic community because it seamlessly integrates with Robotic Operating System (ROS) \cite{takaya2016simulation}, \cite{yao2015simulation}. This allows the users to benefit from Gazebo's 3D visualizations, along with the middleware support from ROS. 

Despite the numerous beneficial features of Gazebo, it is not the most convenient tool to use. The construction of a 3D scene is a manual process. One must collect models of various objects and include them in a world file one by one. The time and effort spent on creating a 3D world grows with the complexity of the 3D scene to be reconstructed and number of models in it. A few researchers have attempted to automate the process. These approaches are discussed below.

\subsubsection{Height-map based automatic 3D world construction in Gazebo}
\cite{lavrenov2017tool} presents a method to automate the process of constructing a 3D virtual environment (world) in Gazebo from a 2D map that is built using real sensor data. The map is represented in the form of an occupancy grid, where a white pixel represents a free space, black an occupied space and gray represents a space that is not clearly defined. Once the map is obtained, it is processed through two intermediate steps before a world file is generated for Gazebo.

In the first step, the occupancy grid map is filtered using a modified non-linear filter to remove noise. Then, the filtered map is converted into a grayscale image to prepare for the second step. In the second step, the authors use heightmap element of the SDF \cite{sdf}, where the size and height for the 2D space defined in an occupancy map is defined. The heightmap is simply a 2D array where each element of an array represents a height of a certain region in a 2D plane. The resulting virtual environment has texture, height and supports collision. Finally, this world file is launched in Gazebo simulation. The authors validated the efficacy of the method by testing various robot models in virtual environments that were constructed using the defined method.

\subsubsection{LIRS World Construction Tool (LIRS-WCT)}
In \cite{abbyasov2020automatic}, authors present a tool called LIRS-World Construction Tool (LIRS-WCT) that automates the process of generating a world file for Gazebo from 2D images or a 2D laser range data. 

The conversion process occurs in three steps. The first step converts a grey scale image to a 3D model. This is done by first transforming a 2D image into a heightmap and then converting the heightmap into a COLLADA [53] format, which is one of the mesh formats supported by Gazebo. Then in the second step, texture is applied to the mesh where it gets the visual features. Finally, during the third step of the procedure, a world file is generated which is used to launch the model in Gazebo that was created in steps 1 and 2.

Abbyasov et al. \cite{abbyasov2020automatic} validated the efficacy of the tool by using it to create virtual environments in Gazebo and using them to test various navigation algorithms. The authors mention that the LIRS-WCT is more efficient than the approach in \cite{lavrenov2017tool} because of its added features such as scalability and it’s relatively higher performance measured in terms of the real time factor in Gazebo.

\subsubsection{map2gazebo}
The map2gazebo is an open source ROS package \cite{gazebopack} that creates a Gazebo world from a 2D map. A ROS node called map2gazebo subscribes to a topic where the static map is being published and it creates a mesh from a static 2D map. The 3D mesh generated from a 2D map is made by extruding occupied pixels in the occupancy grid map. Thus, the 3D Gazebo world generated is basically a skeleton world with boxes without any texture. This may be sufficient to test basic object avoidance models but if a more complicated environment is desired, then the user must re-work the created world by adding other objects to it. 

\subsubsection{USARSim}
The Urban Search and Rescue Simulation (USARSim) \cite{carpin2007usarsim} is an open source, extensible, multi-robot, and flexible simulator which was originally developed for urban and rescue type of robotic applications. It builds on an Unreal gaming engine and supports all major operating systems. It is widely used during the RoboCup competitions. There is also an interface that connects USARSim with ROS called USARSimROS \cite{balakirsky2012usarsim}, however, it does not have a strong community support like Gazebo and is not as well maintained as Gazebo.

\subsection{Middleware for Robotics Development}
The robotic middleware facilitates faster, collaborative, and efficient development of robotic systems. The simulators discussed in the previous section are often used in conjunction with the middleware frameworks. The Robotic Operating System (ROS) is one the most popular middleware, however, few other software frameworks that are similar to ROS also exist. We briefly introduce them in this section. 

\subsubsection{Robotic Operating System (ROS)}
It is an open source, robotic middleware that supports multiple languages such as Python, Octave, C++ and LISP \cite{quigley2009ros}, \cite{joseph2018robot}. It’s modular design approach makes it suitable for use in peer to peer systems where multiple processes implemented on heterogeneous systems are able to communicate with each other. There are four basic elements in ROS architecture, namely the nodes, topics, messages and services. The node is essentially a software process that is designed to do a specific task. A ROS based system usually has many specialized nodes that work together. Each node is capable of publishing and subscribing data to and from other nodes through channels called topics. These topics contain messages. The design choice of dividing the work amongst various specialized nodes makes it easier to debug and it enables a collaborative development.

ROS also offers logging and playback capability, thus making it easier to run offline simulations with the real data.  Additionally, it supports visualization through a program called rviz. However, it is often used in conjunction with other 3D simulators such as Gazebo for enhanced visualizations \cite{takaya2016simulation}, \cite{yao2015simulation}. Finally, like Gazebo, ROS also has a huge community that actively supports and contributes to the open source middleware. The ROS community has released ROS2 recently.

\subsubsection{LCM: Lightweight Communications and Marshalling}
LCM [38] is a library designed to facilitate communication between various robotic processes in real-time systems. LCM supports message definition that is language agnostic. The messages are passed based on the publish/subscribe mechanism which is based on the UDP multicast. There is no central master in LCM unlike in ROS. \cite{quigley2009ros} describes in detail how message mapping is achieved in the absence of a central hub. The notable feature of LCM is that it also has a tool to debug and inspect messages without any significant toll on the efficiency and performance of the system. 

It is important to note that LCM is not a full-fledged middleware like ROS. It is mainly an enhanced message passing library that can be integrated with the existing software framework. 
\subsubsection{ZeroMQ}
This is yet another open source and popular library for communication purposes \cite{sustrik2015zeromq}.  It supports different types of message models for message delivery such as a centralized approach where there is a central hub responsible for coordinating message flow between various constituent processes, a peer to peer coupling, a method where a central hub is used as a directory server and also a distributed system that can have one or many central hubs.

\subsection{Simulators for human behavior modelling}
\label{ped_modelling}
The simulators discussed in this section are called pedestrian simulators. Their aim is to facilitate research in studying crowd behavior such as passenger flow at the airports, railway stations, or for urban planning. There are several commercial and few open source pedestrian simulators available in the market. The basic building blocks of these simulators are the mathematical pedestrian models, such as the social force model \cite{caramuta2017survey}, \cite{helbing1995social}, which simulate how pedestrians generally move in a shared environment. While these simulators are quite mature and are widely used in testing new pedestrian models, most of them do not have a support for testing mobile robots in the same simulation environment. This section reviews most popular pedestrian simulators. 

\subsubsection{Menge}
It is an open source, cross platform and extensible simulator designed to facilitate research in pedestrian crowd dynamics \cite{curtis2016menge}. It is based on microscopic models of pedestrians where each of the human models is considered as an independent agent moving in a shared space. The problem of simulating crowd dynamics is divided into several sub-problems, which are then individually addressed. This makes Menge a modular and flexible framework.

The pipeline of various sub-problems include goal selection, path computation and path adaptation.  The plan adaptation step uses various pedestrian models to compute a more realistic velocity called the feasible velocity which aims to prevent collision with the fellow pedestrians in the environment. Menge comes pre-loaded with various pedestrian path prediction models built into it. These include PedVO \cite{van2011reciprocal}, helbing \cite{helbing1995social}, karamouzas \cite{karamouzas2009predictive}, johansson \cite{helbing2007dynamics}. It also includes ORCA \cite{van2011reciprocal} as the collision avoidance model. Besides the elements that are already part of the Menge library, it allows users to create their own algorithms and plug into Menge’s architecture for the purpose of evaluating their algorithms.

Menge is a versatile simulator for studying crowd dynamics, however, it is not suited for testing mobile robots without the additional work. 


\subsubsection{PEDSIM}
PEDSIM [45] is an open source simulator that is based on the social force model \cite{helbing1995social} and the extended social force model. This is mainly used to study crowd behavior of pedestrians in both indoor and outdoor environments. It is based on a standalone C++ library called libpedsim, which is easy to extend. Typically, the pedestrian trajectories calculated by PEDSIM are exported and visualized in any rendering engine of user’s choice. It is capable of running on both Linux and Windows operating systems. The key feature of PEDSIM is that is easy to use and can be extended with less effort.

There are several other pedestrian simulators, such Mass Motion, Pedestrian Dynamics, Legion, Simtread, Simwalk, Anylogic, Viswalk, Simped, and Alpsim. These are reveiwed in  \cite{caramuta2017survey} and \cite{kielar2016momentumv2}.

\subsection{Simulators for combined robot and human modelling}
\label{social_mobile_sim}
The simulators for testing mobile social robots are unique as they must support all of the key requirements mentioned in Section \ref{requirements}. This section elaborates on the current state-of-the-art simulators that aim at a facilitating simulation testing for social mobile robots in pedestrian rich environments.

\subsubsection{MengeROS} \cite{aroor2018mengeros} extends Menge’s functionality by integrating Menge\cite{curtis2016menge} with ROS\cite{quigley2009ros} in order to provide a simulation tool for testing mobile robots in pedestrian rich environments. MengeROS is an open source ROS package. It is implemented as a single ROS node where the pedestrians and mobile robots are both present. This node takes in velocity as an input that moves the robot and provides laser readings of the environment as an output. The robot avoids collisions with pedestrians that are moving per Menge’s specifications after processing the laser sensor readings. 

It is important to note that MengeROS is a 2D simulator where the pedestrians and mobile robots are represented as circles. This type of simulation framework may suffice for testing general robot path planning purposes. If a researcher is interested in gesture recognition, face detection or other distinctive features of pedestrians, MengeROS is not enough. 

\subsubsection{menge\_core}
\cite{mengecore} is a work that is in experimental stages. It provides an open source library that integrates Menge \cite{curtis2016menge} with Gazebo \cite{koenig2004design} simulator. This library serves as a crowd simulator that supports mobile robots. The author introduces the concept of an external agent, which separates it from rest of the simulation moving objects such as pedestrians. One of the important features of this library is that it can represent robots and pedestrians as 3D objects. This library is in developmental stages at the time of this writing.

\subsubsection{Pedsim\_ros}
Pedsim\_ros \cite{pedsimros} is a ROS package that adapts PEDSIM simulator to be used in conjunction with ROS \cite{quigley2009ros}. It offers all the benefits of PEDSIM simulator while also supporting robot integration with ROS. It also supports realistic visualization of human models in Gazebo \cite{koenig2004design}. Pedsim\_ros is used in \cite{du2019group} to evaluate a navigation model for sidewalk robot navigation. Additionally, \cite{holtz2021socialgym} uses Pedsim\_ros to model human behavior for the proposed SOCIALGYM 2D simulation framework.

While Pedsim\_ros certainly meets all the key requirements for a pedestrian simulator mentioned in Section \ref{requirements}, it is important to note that in order to use this package, the user must create a 3D virtual environment and human models in Gazebo from scratch. A second drawback is that Pedsim\_ros supports a single mathematical model for pedestrians unlike Menge \cite{curtis2016menge} and MengeROS \cite{aroor2018mengeros}.

\subsubsection{SocNavBench}
SocNavBnech framework \cite{biswas2021socnavbench} is the most recent advancement in this area and as such it takes a very unique approach to modelling pedestrians in simulation. It uses real world pedestrian data to simulate human behavior as opposed to relying on mathematical models, which are the backbone of all the simulators (Sections \ref{ped_modelling} and \ref{social_mobile_sim}) built for this task so far. Additionally, the simulation framework comes with a default mobile robot model. Furthermore, the authors re-create virtual environments that correspond to the actual physical locations where pedestrian data was originally recorded. The simulator has two modes of operation, schematic and full render. The full render mode provides photogenic/3D view of the physical environment and pedestrians. This type of framework allows one to have a more realistic representation of a physical environment to test navigation algorithms for mobile robots.  As the pedestrian activities are rooted in the real world data, it naturally contains the diversity of activities that pedestrians are known to perform in public, such as walking, running, crossing, grouping, etc. We believe SocNavBench is a significant advancement in this area. However, it has few limitations that we discuss in Sections \ref{evaluation} and \ref{challenges}. 

In Section \ref{social_mobile_sim}, we described the current state-of-the-art simulation frameworks for testing mobile robots in pedestrian rich environments. We want to highlight that besides these simulation frameworks, many researchers have leveraged Gazebo simulator to create their own simulation environments. Some of the notable works include 
\cite{he2020method} which introduces a method to integrate human animation models developed using the Editor for Manual Work (EMA) tool into Gazebo simulation. This framework extends the capabilities of Gazebo, while retaining the other key features such as its ability to be integrated with ROS. Additionally, \cite{tai2018socially} releases an open source plugin for simulating humans in Gazebo environment. This plugin models human behavior as per the social force model \cite{helbing1995social}. Furthermore, \cite{liang2021crowd} also uses Gazebo to construct a simulator to test robot navigation in dense pedestrian crowds.

\section{Evaluation}
\label{evaluation}
The simulators in Section  \ref{3D_env_modelling} specialize in simulating 3D virtual environments, robot models and sensor models. Further, Section \ref{ped_modelling} describes the simulators that specialize in modelling pedestrian behavior. However, the simulators for testing navigation algorithms for mobile robots in pedestrian rich environments must have the combined capabilities of these two types of simulators. Section \ref{social_mobile_sim} sheds light on such simulators. We experimented with some of the open source and the commonly used simulators mentioned above. In this section, we provide our analysis as to which simulation frameworks are best for creating 3D environments, robot and human models individually. Then, we discuss which of them are suitable for testing AMRs in pedestrian rich environments. 

\begin{figure}[!t]
\centering
\includegraphics[width=2.5in]{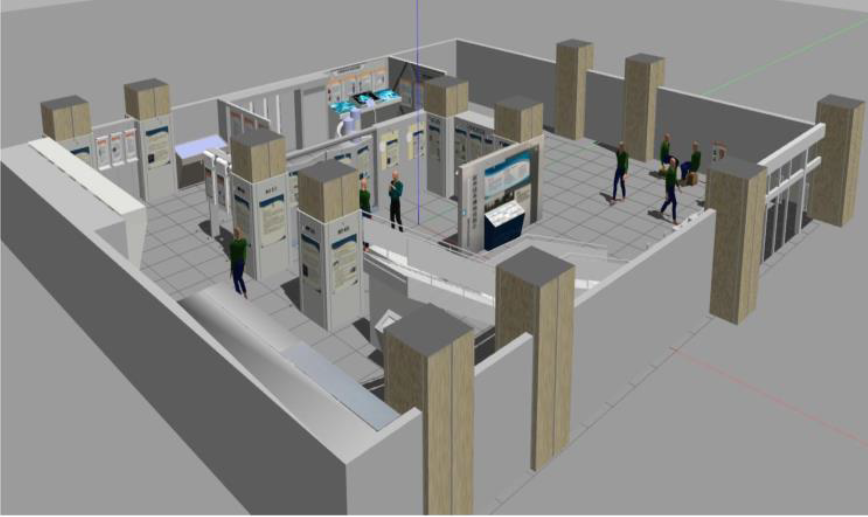}
\caption{A 3D scene of a museum in Gazebo.}
\label{3D_env_fig}
\end{figure}

\begin{figure}[!t]
\centering
\includegraphics[width=2.5in]{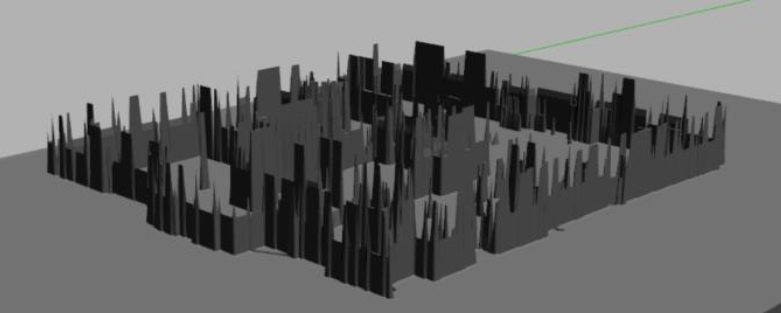}
\caption{A 3D model of the scene in Fig.3 reconstructed using LIRS-WCT.}
\label{LIRS_3D_Fig}
\end{figure}

\begin{figure}[!t]
\centering
\includegraphics[width=2.5in]{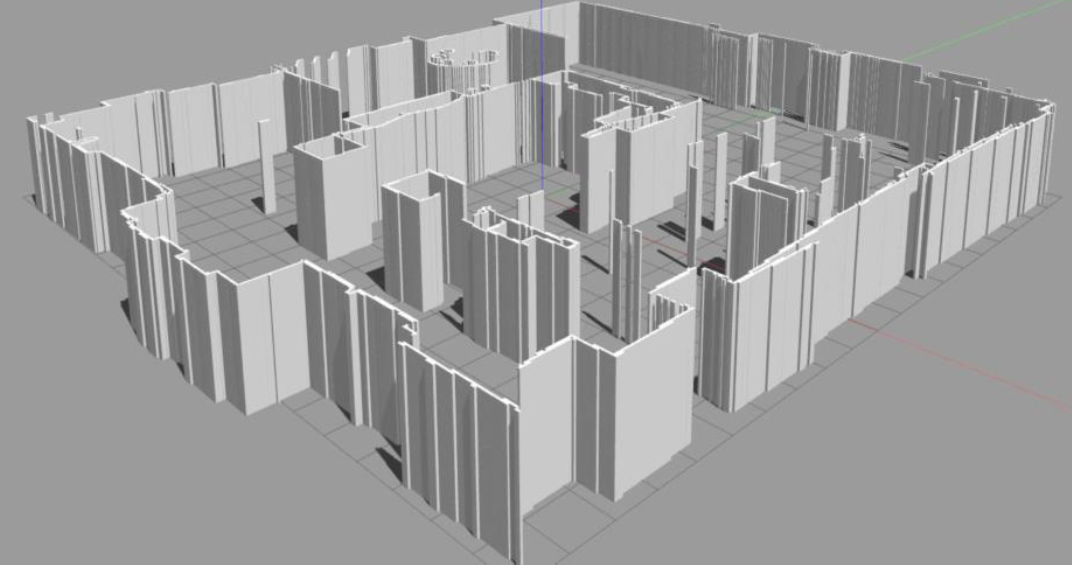}
\caption{A 3D model of the scene in Fig.3 reconstructed using map2gazebo.}
\label{map2gazebo_fig}
\end{figure}

\begin{figure}[!t]
\centering
\includegraphics[width=2.75in]{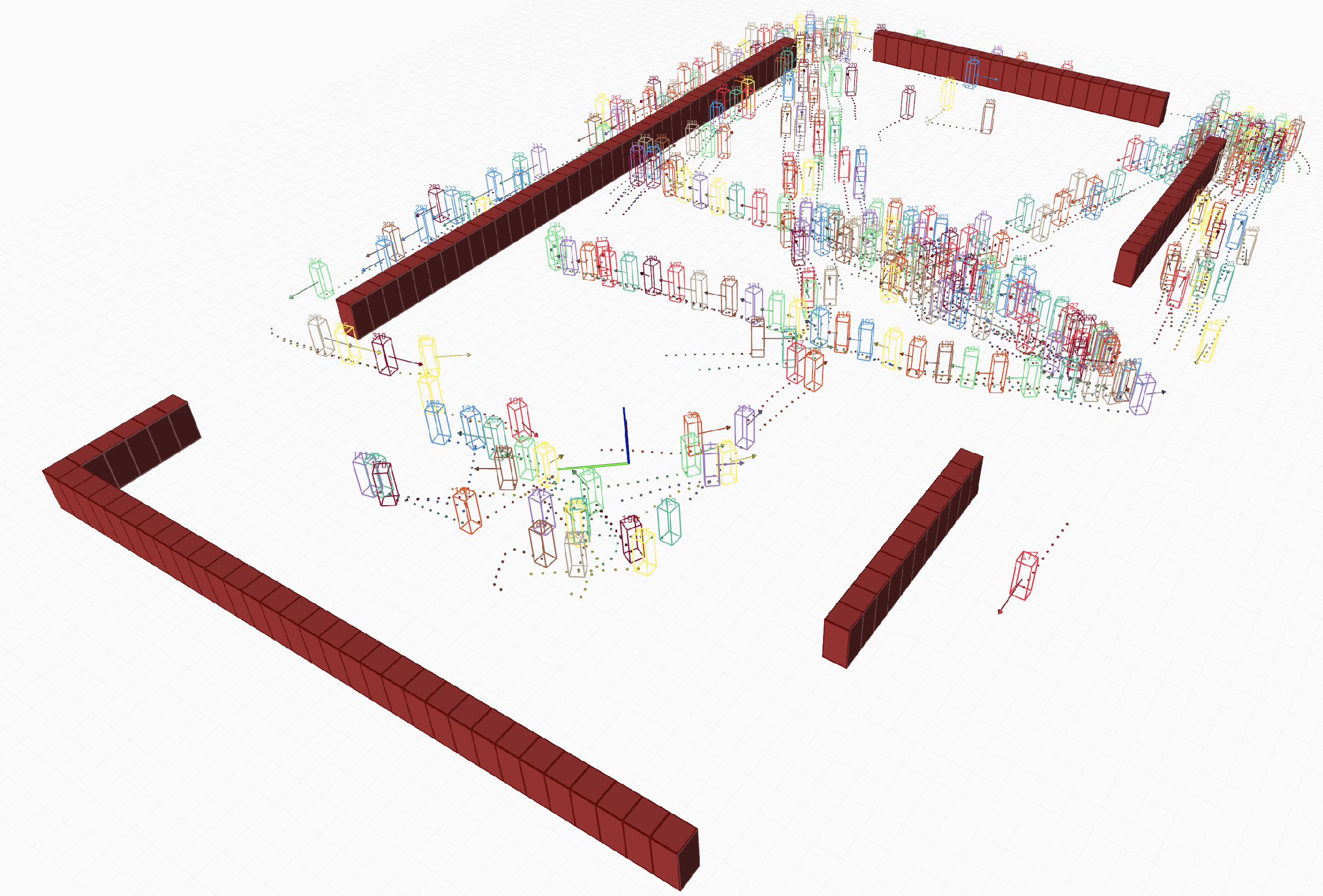}
\caption{3D simulation of pedestrians moving in an airport scenario using the pedsim\_ros package.}
\label{airport_activities}
\end{figure}

\begin{figure}[!t]
\centering
\includegraphics[width=3.5in]{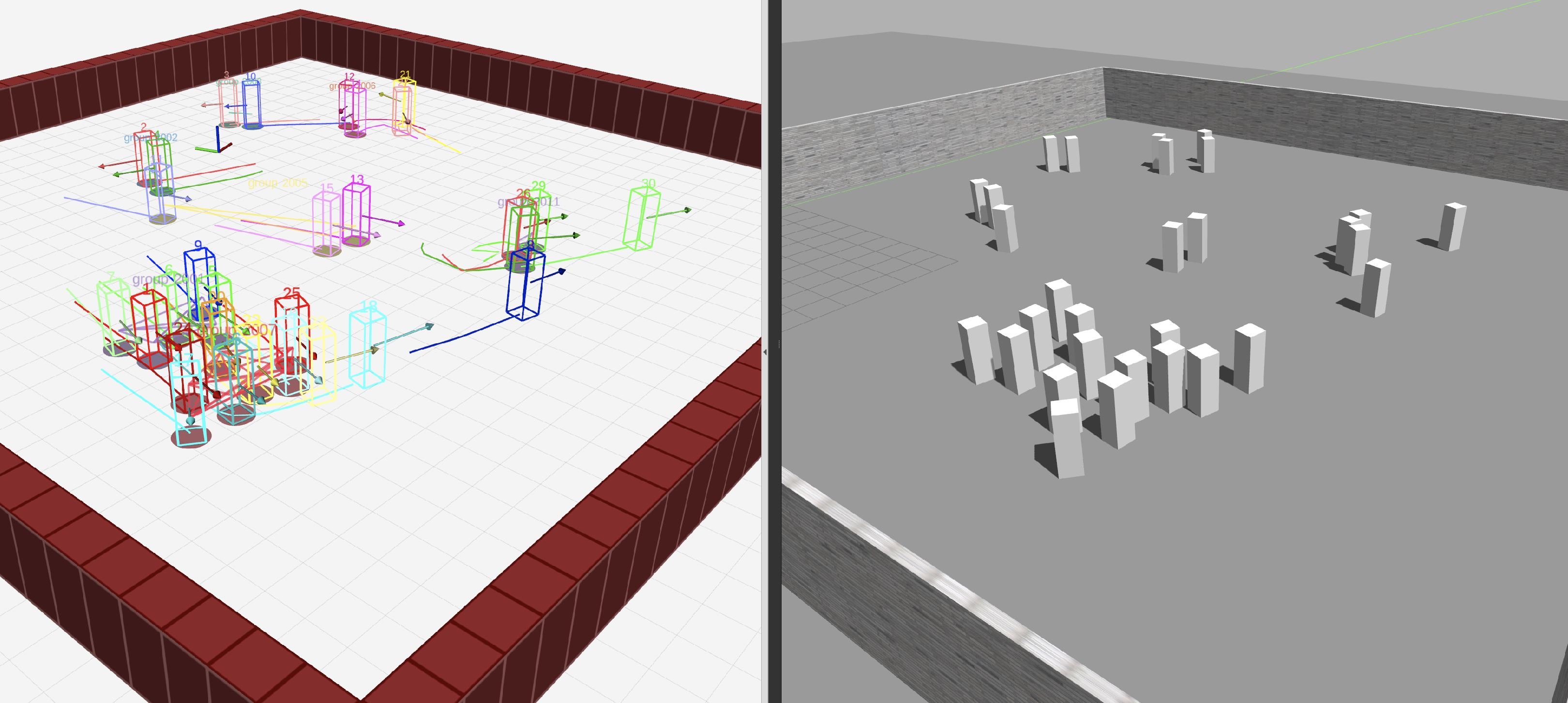}
\caption{3D simulation of pedestrians in rviz and Gazebo using the pedsim\_ros package. The tall boxes in the image represent human models.}
\label{simple_rviz_gazebo_fig}
\end{figure}

\begin{figure}[!t]
\centering
\includegraphics[width=1.5in]{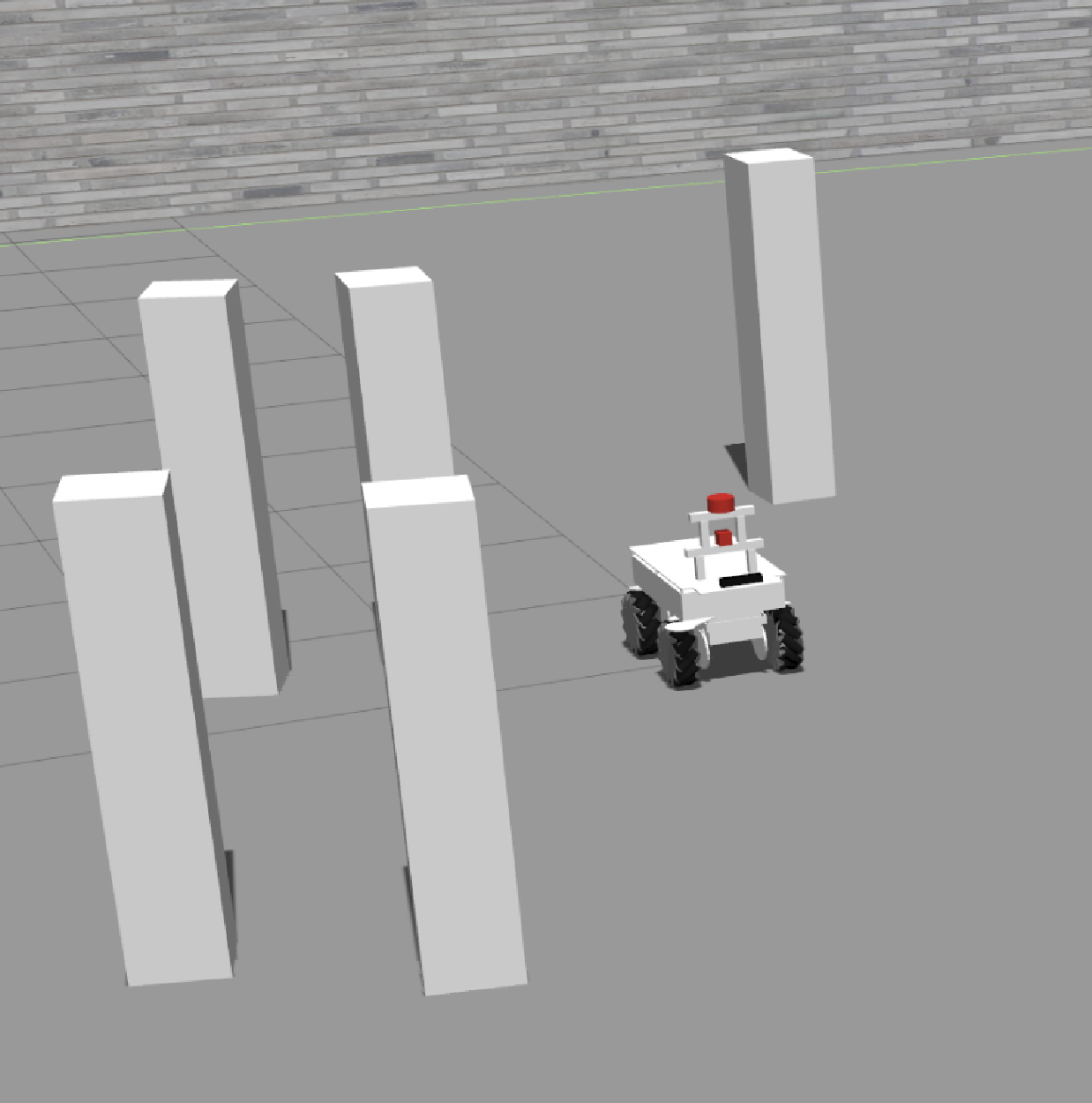}
\caption{3D simulation of pedestrians (tall boxes) along with a custom designed 3D model of a robot in Gazebo.}
\label{simple_with_robot}
\end{figure}

\begin{enumerate}
    \item For creating 3D virtual environments and various kinds of robot models, Gazebo is the most preferred robotic simulator \cite{ivaldi2014tools}  due to its flexibility and easy integration with ROS. The creation of 3D environment is a manual process in Gazebo where each element of the environment is hand crafted. This allows one to simulate different types of road terrains such as the ones with different slopes, indoor versus outdoor environments, etc. While this offers a lot of flexibility to control every aspect of the simulation, creation of such an environment is a time consuming task. Fig. \ref{3D_env_fig} shows an example of a 3D environment of a museum constructed in Gazebo.
    \item To reduce the time consumption involved in creating 3D environments in Gazebo, two open source packages, namely, LIRS-WCT \cite{abbyasov2020automatic} and map2gazebo \cite{gazebopack} have been released recently. These packages aim to automate the creation of 3D environments in Gazebo. We tested both of these frameworks by re-creating a 3D scene shown in Fig. \ref{3D_env_fig}. The results after automatic creation of this scene using LIRS-WCT and map2gazebo are shown in Fig. \ref{LIRS_3D_Fig} and \ref{map2gazebo_fig}, respectively. We believe that map2gazebo package is easier to use and it also gives better results. While these two packages do ease the work load for creating a 3D scene, they require one to have a 2D map of the environment that needs to be fed as an input. So, if one does not have a 2D map of the environment, then these packages do not help. 
    \item For robotic simulation, middleware plays a key role in facilitating lower level communication between various processes. ROS is the preferred middleware in robotics. It is well maintained and can handle complex robotic systems. ROS does come with its own visualization tool called rviz, however, it is often used in conjunction with Gazebo simulator to provide more realistic 3D simulations. 
    \item Finally, for testing AMRs in pedestrian rich environments, some of the traditional pedestrian simulators described in Section \ref{ped_modelling} have been extended by the robotics community for further use. MengeROS \cite{aroor2018mengeros} and Pedsim\_ros are two such simulators. These have been released as open source ROS packages to test AMRs in pedestrian rich environments. MengeROS and Pedsim\_ros use traditional pedestrian simulators Menge and PEDSIM, respectively, to model pedestrian behavior, while relying on Gazebo and ROS for modelling robots and their sensors in a 3D virtual environment. Additionally, SocNavBench is a recent add to the community. We attempted to test both MengeROS (a 2D simulator) and Pedsim\_ros (a 3D simulator). While MengeROS was not testable as it is way behind the current ROS release. Additionally, this package does not appear it is being actively maintained. However, we were able to successfully test the performance of Pedsim\_ros. Next, we provide a summary of its capabilities and performance.\\
    
    As the name suggests, Pedsim\_ros integrates PEDSIM simulator with ROS. Additionally, it also has a plugin to connect with Gazebo. This gives user a powerful combination that meets all of the key requirements mentioned in Section \ref{requirements} above. Fig. \ref{airport_activities} shows an example of an airport environment in rviz where pedestrians are moving in crowds. The pedestrian behavior is as per the mathematical model called the social force model \cite{helbing1995social}, that is pre-programmed in the PEDSIM simulator. Therefore, if researchers are interested in designing and testing a mobile robot in such an environment, they can focus on a specific task such as designing a robot that maintains a collision free path without having to model the pedestrian behavior themselves manually. Further, to evaluate the Gazebo plugin, we can see the comparison of the 3D scene with pedestrians in rviz and Gazebo simultaneously in Fig. \ref{simple_rviz_gazebo_fig}. (Note: The tall boxes in the simulation correspond to human models.) Finally, in Fig. \ref{simple_with_robot}, we show a 3D model of a robot along with the pedestrian models (tall boxes) in Gazebo. The robot model shown is constructed in Gazebo from scratch. We believe that Pedsim\_ros is a great start for testing AMRs in pedestrian environments as it allows users to focus on modeling robot behavior while it takes care of the pedestrian modelling itself. This is especially true if the robot path planning and control algorithms solely rely on predicted pedestrian trajectories. However, if a navigation algorithm relies other types of human robot interactions such as human body orientation, skeleton tracking, gestures or gaze estimation \cite{9636816}, then Pedsim\_ros is not an ideal choice.  
    
    To overcome the shortcomings of existing simulators, SocNavBench \cite{biswas2021socnavbench}, a very recent add to the simulation frameworks in this area, offers added features such as 3D rendering of human figures. The human meshes used in the simulation represent humans of different sizes and shapes, performing diverse activities. Additionally, the human trajectories are derived from the real world data, which makes the scene more representative of a real environment. The current shortcomings of SocNavBench are that it is not integrated with ROS and does not support creation of robot models and sensor models from scratch as in Gazebo. We believe that ability to integrate with ROS is critical as ROS is the most widely used middleware in the robotics community. Additionally, creation of robot and sensor models from scratch is also essential as robots come in different forms, equipped with wide variety of sensors. It is pertinent to be able to test the kinematics and sensor performance for a specific robot type in a simulated environment.

\end{enumerate}

See Table 1 for comparison of various simulators reviewed as part of our survey.

\begin{table}[!t]
\caption{A Comparison of simulation frameworks}
\label{table_example}
\centering
\begin{tabular}{|p{0.150\linewidth}|p{0.125\linewidth}|p{0.125\linewidth}|p{0.125\linewidth}|p{0.125\linewidth}|p{0.125\linewidth}|}
\hline
 \textbf{Simulator}& \textbf{3D environment modelling} & \textbf{Robot modelling} & \textbf{Sensor modelling} & \textbf{Integration with robotic middleware}& \textbf{Human modelling}\\
\hline
\textbf{Gazebo/ Ignition} & Y &Y &Y& Y, with ROS &Y, limited ability \\
\hline
\textbf{USARSim} & Y & Y & N&Y, with ROS&N\\
\hline
\textbf{Menge} & N & N&N &N&Y, limited to 2D\\
\hline
\textbf{PEDSIM} & N & N &N &N&Y\\
\hline
\textbf{MengeROS} & N &Y, limited to 2D& N&Y, with ROS&Y,limited to 2D\\
\hline
\textbf{Pedsim\_ros} & Y, with Gazebo interface &Y, with Gazebo interface &Y, with Gazebo interface&Y, with ROS&Y, supports 3D\\
\hline
\textbf{SocNavBench} & Y&N, supports a default model &N &N &Y \\
\hline
\multicolumn{4}{l}{$^{\mathrm{a}}$Y=Yes, N=No.}
\end{tabular}
\end{table}

\section{Challenges}
\label{challenges}
\begin{figure}[!t]
\centering
\includegraphics[width=3.0in]{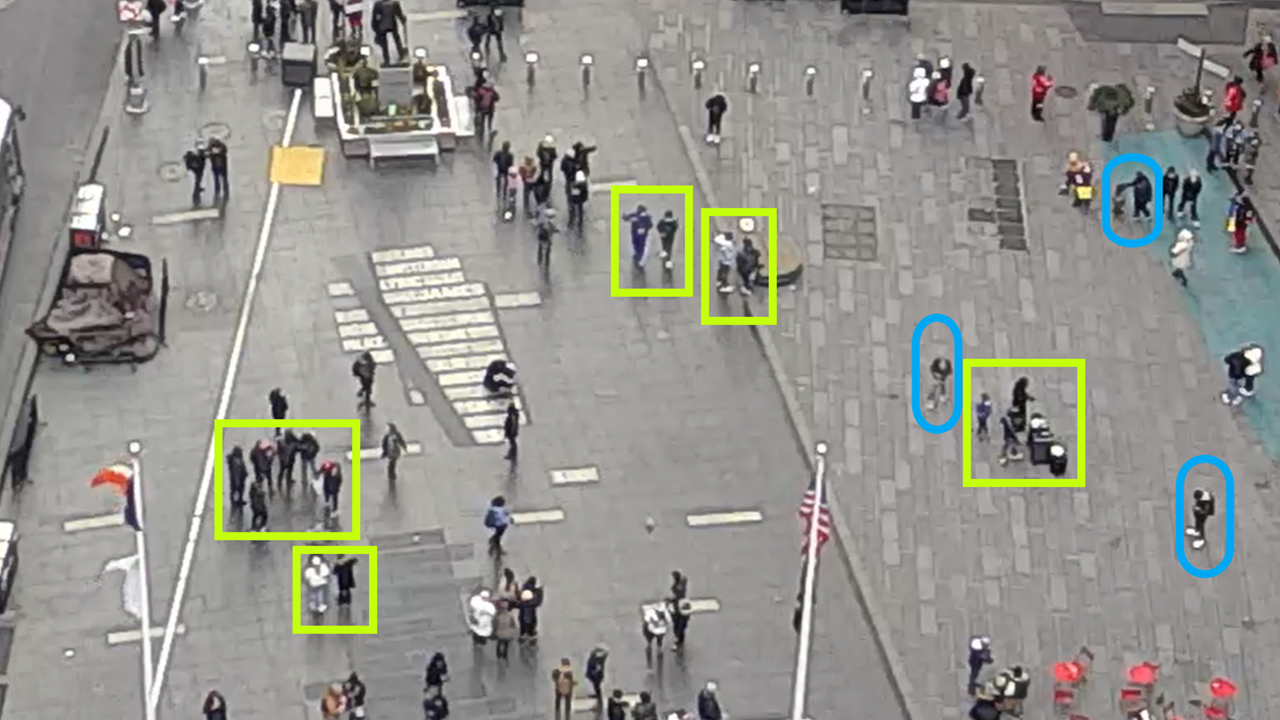}
\caption{Live footage of pedestrians at Times Square, NYC as captured by EarthCam. The annotated rectangular boxes and ovals represent groups of pedestrians and individuals that do not belong to any group, respectively.}
\label{fig_sim2}
\end{figure}

This section discusses challenges and limitations  of the existing simulators.
\begin{enumerate}
    \item \textit{Inability to dynamically switch between pedestrian models:} All pedestrian simulators except the SocNavBench framework rely on fixed mathematical models (either, microscopic or macroscopic) to simulate pedestrian behavior. This implies that the simulators model pedestrian dynamics either at the crowd level or at the individual level. However, the amount of information required to asses a potential human-robot collision at any time is highly dependent on the distance between them at that time. Therefore, it is necessary to model pedestrians using both, the macroscopic and macroscopic methods, depending on the immediate situation around the robot. As an example, when the human crowd is at a farther distance from the robot, it is sufficient to use macroscopic mathematical models which define the properties of the crowd such as its velocity and position. This would give enough information to guide robot's next steps as shown in \cite{wang2022group}. However, as the crowd gets closer to the robot in terms of its distance, the microscopic pedestrian models are more suitable to provide the details of each pedestrian's dynamics. Currently, no simulator supports dynamic switching between macroscopic and microscopic pedestrian models as a response to the changing environment around the robot. SocialNavBench framework has an added advantage here as it uses real pedestrian data to model pedestrian trajectories. Threfore, the different types of pedestrian dynamics, such as the corwd behavior versus an individual behavior are intrinsically built in the simulated environment. 
    \item \textit{Absence of modelling of human behavior in the presence of a robot:} All simulators in Section \ref{ped_modelling} and \ref{social_mobile_sim} have a common limitation. While these models are capable of modelling human interactions at the individual level and at the group level, they do not incorporate human behavior in the presence of a robot. The human behavior changes during the presence of a robot due to its novelty \cite{kidokoro2015simulation}, \cite{chen2019crowd}. No simulator currently models such interactions. This limits robot’s ability to plan its path in a real environment where people are expected to form crowds around a robot. 
    \item \textit{Lack of diverse pedestrian models:} All simulators except SocNavBench suffer from homogeneity as they deploy one fixed mathematical model for all pedestrians in the scene. Since the behavior of humans differ depending upon their age, demographics, gender and culture \cite{hurtado2021learning}, \cite{gao2022evaluation}, \cite{mavrogiannis2021core}, a single pedestrian model applied to all human models is not a realistic representation of human population, especially in the public places like hospitals, airports, hotels, etc. SocNavBench is an exception as it is rooted in the real pedestrian data for simulating pedestrian behavior.
    \item \textit{Absence of modelling of robot to robot interactions:} It is expected for public places such as hospitals, nursing homes, airports, and construction sites to have multiple robots from different manufacturers working together. In order for them to work efficiently, they must be able to communicate with each other. An example of issues where robots from different vendors came to a halt because they did not know how to respond to each other is recently reported from the Changi General Hospital in Singapore, which has about 50 robots. Therefore, testing a stand alone AMR in an environment is not sufficient, especially if it is intended for a site that is likely to have robots from other manufacturers in its vicinity.
    \item \textit{Absence of various visual cues for navigation such as human gaze/head/body orientation: } Some robotic algorithms rely on visual cues \cite{tolani2021visual} such as human gaze or head and body orientation to do the path planning \cite{chen2020robot}, \cite{singh2018combining}. The current simulators do not support testing such algorithms because they either model pedestrians as 2D, or 3D human models with minimal to no facial features. The importance of such navigation cues is evident from the the live images of pedestrians shown in Fig. \ref{fig_sim1} and Fig. \ref{fig_sim2}. These images are from Times Square in New York as captured by the EarthCam network. After careful observation, a human can easily distinguish various different social groups as well as the individuals that do not belong to any group. Such distinction is possible by examining the type of interaction occurring between individuals and other cues such as the orientation of their head/body. This type of granularity is currently lacking in the simulators we have today. 
    \item \textit{Absence of ambient sound modelling in the virtual scene: }Additionally, some robots use environmental sound or human speech for navigation \cite{nakadai2000active},  \cite{liu2010continuous}, \cite{rascon2017localization}, and \cite{9040915}. Currently, there is no simulator for testing sound based navigation algorithms..
    \item \textit{Lack of realistic noise modelling in simulation:} Although some of the simulators such as Gazebo allow one to model sensor noise for various sensors such as for lidar and camera \cite{furrer2016rotors}, \cite{lavrenov2017tool}, there are many types of noises that currently cannot be modelled in simulation. This includes ambient noise for the acoustic input based navigation algorithms \cite{martinson2004noise} or a noise originating from the robot itself through its motors and fans \cite{martinson2007hiding}. A proper amount of noise modelling in the simulation environment will help fill the current gap between real and the simulated environments \cite{liang2021crowd}.
    \item \textit{Lack of automatic 3D scene generation: } \cite{afzal2020study} presents ten challenges of using simulators that roboticists currently face based on the survey that involved 82 roboticists from various backgrounds. The data shows that since the construction of a simulation environment with the current methods is a tedious and time-consuming task, the researchers hesitate using simulators all together. In order to enhance simulation testing and make it more efficient, there is a need for more open source models of various 3D environments. This will help the entire robotic community to design complex scenarios without re-inventing the wheel every time.
    
\begin{figure}[!t]
\centering
\includegraphics[width=3.0in]{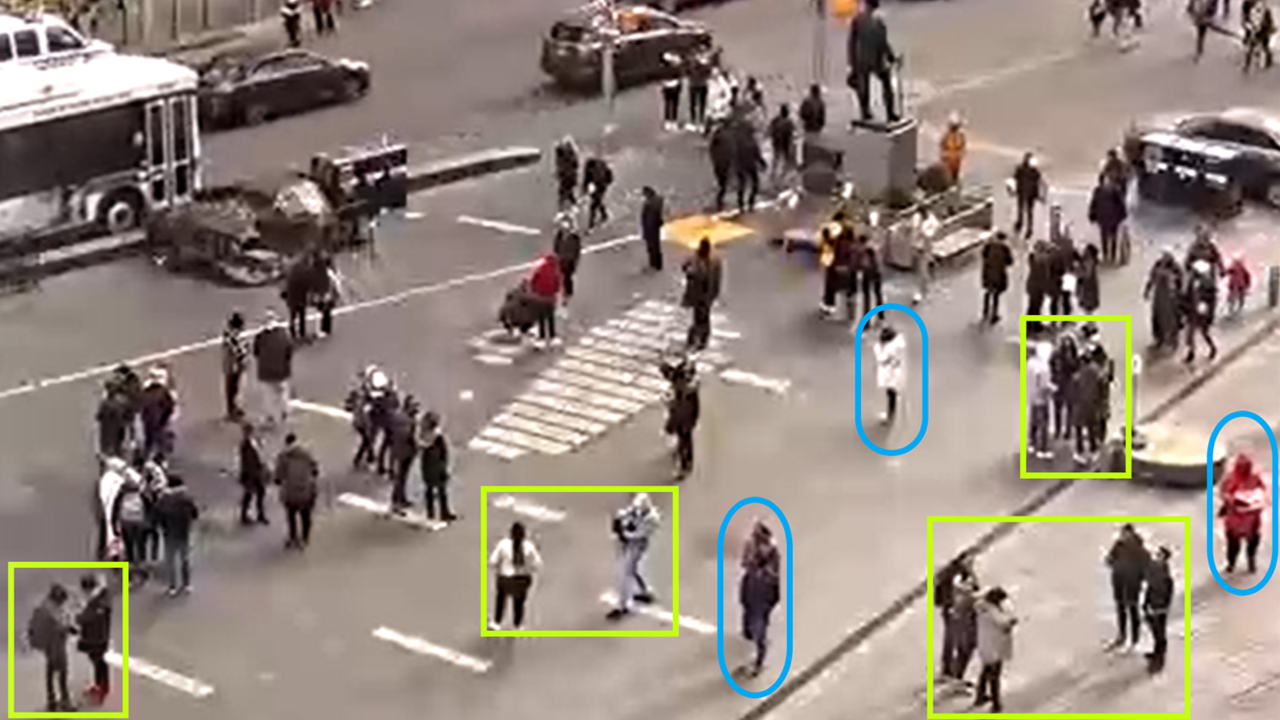}
\caption{Live footage of pedestrians at Times Square, NYC as captured by EarthCam. The annotated rectangular boxes and ovals represent groups of pedestrians and individuals that do not belong to any group, respectively.}
\label{fig_sim1}
\end{figure}

\end{enumerate}

\section{Related Works}
There are few other 3D simulators like Gazebo that are presented in \cite{koenig2004design}, \cite{staranowicz2011survey}, \cite{torres2016survey} and  \cite{castillo2010introductory}. \cite{pitonakova2018feature} also provides a very detailed comparison of Gazebo with other alike simulators such as V-REP and ARGoS. Similarly, a comprehensive comparison of various robotic middleware is provided in \cite{magyar2015comparison}, \cite{quigley2009ros}, and \cite{tsardoulias2017robotic}. \cite{magyar2015comparison} compares RT-Middleware, Robotic Operating System (ROS), Open Software Platform for Robotics Services (OPRoS), and Orocos based on their capabilities such as the support for various operating systems, real time operability, programming languages supported and their respective simulation environments. A comparison of Player, ROS, Microsoft Robotics Studio (MSRS), Webots, Robot Construction Kit (ROCK), iRobot Aware, Python Robotics (Pyro), Carmen (Robot Navigation Toolkit) can be found in \cite{tsardoulias2017robotic}. Further, \cite{caramuta2017survey} and \cite{kielar2016momentumv2} provide a survey on various pedestrian simulators. Finally, \cite{tsoi2020sean} introduces a new framework called SEAN-EP which is an experimental platform that allows human in the loop robot navigation.

\section{Conclusion}
\label{sec:conclusion}
This paper reviewed available simulation tools to test social mobile robots in pedestrian rich environments. It is concluded that there exist various simulators such as Gazebo that specialize in testing robot path planning algorithms in static environments. Further, there also exist many pedestrian simulators that specialize in testing pedestrian behavior models. However, there are very few simulation frameworks for testing mobile robots in human environments. According to our study, SocNavBench and Pedsim\_ros are the only 3D frameworks currently suitable for testing social AMRs. Further, the paper identifies several challenges and limitations that the current simulation techniques have. There is a need for more open source simulators that will enable effective testing of AMRs. This can help accelerate the deployment of mobile robots in complex human environments.

\bibliography{bibtex/bib/IEEEabrv.bib,bibtex/bib/survey.bib}{}
\bibliographystyle{IEEEtran}

\end{document}